\documentclass[conference]{IEEEtran}
\usepackage{amsmath}
\usepackage{graphicx}
\usepackage{hyperref}
\usepackage{amsfonts}
\usepackage{cite}

\title{PALMS: Plane-based Accessible Indoor Localization Using Mobile Smartphones}
\author{
    \IEEEauthorblockN{Yunqian Cheng}
    \IEEEauthorblockA{Computer Science and Engineering\\
    University of California, Santa Cruz\\
    Santa Cruz, United States\\
    Email: ychen827@ucsc.edu}
    \and
    \IEEEauthorblockN{Roberto Manduchi}
    \IEEEauthorblockA{Computer Science and Engineering\\
    University of California, Santa Cruz\\
    Santa Cruz, United States\\
    Email: manduchi@ucsc.edu}
}

\begin{document}

\maketitle

\begin{abstract}
In this paper, we present PALMS, an innovative indoor global localization and relocalization system for mobile smartphones that utilizes publicly available floor plans. Unlike most vision-based methods that require constant visual input, our system adopts a dynamic form of localization that considers a single instantaneous observation and odometry data. The core contribution of this work is the introduction of a particle filter initialization method that leverages the Certainly Empty Space (CES) constraint along with principal orientation matching. This approach creates a spatial probability distribution of the device's location, significantly improving localization accuracy and reducing particle filter convergence time. Our experimental evaluations demonstrate that PALMS outperforms traditional methods with uniformly initialized particle filters, providing a more efficient and accessible approach to indoor wayfinding. By eliminating the need for prior environmental fingerprinting, PALMS provides a scalable and practical approach to indoor navigation.
Full code and data can be accessed at the following link: \href{https://github.com/Head-inthe-Cloud/PALMS-Indoor-Localization/tree/main}{https://github.com/Head-inthe-Cloud/PALMS-Indoor-Localization/tree/main}

\end{abstract}


\section{Introduction}
Indoor wayfinding, the ability to determine and follow a path in a building towards a desired destination, is an open area of research, with applications including travel (e.g., finding a gate in an airport), health (e.g., finding a doctor’s office), and recreation (e.g., exploring a museum). Wayfinding is particularly challenging for those who are blind or have low vision, as these individuals cannot rely on visual landmarks for orientation, and cannot easily consult a map of the place.

Different types of smartphone apps for assisted wayfinding are available. Some (like Google Maps and Apple Maps) are general-purpose localization and guidance apps that can work indoors (even when GPS signal is unusable) using radio signal from Wi-Fi access points or Bluetooth Low Energy (BLE) beacons. This requires a prior {\em fingerprinting} phase which, generally speaking, involves collecting signal measurements (received signal strength or RSSI) over a dense grid of spatial locations in the building. Indoor Atlas relies on “magnetic signatures” for mapping and also requires prior fingerprinting of the environment. GoodMaps is a wayfinding app specifically designed for blind users. It enables indoor mapping using images taken from a smartphone camera, which are matched, using computer vision techniques, against a database of visual features collected in a prior phase (what could be termed  {\em visual fingerprinting}).

A disadvantage of these technologies is that localization is only available in environments that have been fingerprinted, which represents a small portion of the available buildings. A more scalable approach, one that requires no fingerprinting, is offered by systems that attempt to localize and track the walker using dead-reckoning. For example, Crabb et al. \cite{Crabb2023ALA} experimented with wayfinding using ARKit, an iOS framework for vision-inertial odometry (VIO). A user of this system walks while holding the smartphone in such a way that the camera has a clear view of the environment. A structure-from-motion algorithm, complemented with inertial data, measures the phone’s velocity and orientation with respect to a fixed reference frame. By integrating the velocity vectors through time, the walker’s trajectory can be reconstructed. The unavoidable drift (due to instantaneous bias and noise) can be compensated for via particle filtering, which uses the floor plan of the building to condition the reconstructed trajectory (in particular, by discouraging trajectories going through a wall). A similar approach was taken by Ren et al.\cite{Ren2021SmartphoneBasedIO, Tsai2024AllTW} but relying on inertial sensors only, using step-based pedestrian dead reckoning (PDR) or machine learning-based odometry (RoNIN).

A main problem with dead-reckoning odometry is that the starting location and orientation needs to be known, requiring some form of initial localization. In addition, occasional relocalization may be necessary because of sporadic system fault or poor tracking due to accumulated drift. In some cases, recognizable visual landmarks, (e.g., an EXIT sign) whose location in the building is known, can be used for initial localization or relocalization (in the following, we will use the term {\em localization} for both cases). However, map annotations of this type are not often available.

In this work, we will only consider prior information about the environment in the form of a floor plan of the building. Floor plans are often available, especially for public places, where they are frequently posted online. They normally represent structural elements (e.g. walls) that, unlike furniture or other objects, are expected to remain stable in time. Floor plan-based localization is highly desirable as it removes the need for environment fingerprinting or detailed map annotations. 
This task is similar to that of a visitor of a building trying to locate themselves within a posted map when the “You are here” marking is missing. 
One may, for example, search the floor plan for a location where the presence of nearby features (e.g., doors, wall corners, hallway size) and their spatial relation, as indicated in the map, conform with the visible scene at their location. Our proposed PALMS (Plane-based Accessible Indoor Localization Using Mobile Smartphones) system uses 3-D data, acquired by an iPhone LiDAR, to detect specific features of the visible wall geometry which are then matched against the wall structure shown in the floor plan.

Regardless of the methodology considered, it is reasonable to expect that, in a building with self-repeating structural elements (e.g., a long corridor with uniformly spaced doors), for each location there may be multiple other locations from which the visual scene looks similar. To mitigate this ambiguity, we  consider a dynamic form of localization: users take an image (or, in our case, a 3-D scan) of the environment from their location; then, they start walking in any available direction, following a certain path (e.g. random). Over a (hopefully short) period of time, the system determines their location and orientation, by combining information from the initial data collection along with dead-reckoning odometry information from the sensor data, conditioned by a particle filter. Intuitively, while multiple locations in the floor plan may be consistent with the visual data recorded at a given location, the same path (from dead-reckoning) is likely to be unfeasible (because of physical constraints) if started from an incorrect location. Inspired by \cite{Ito2014WRGBDFI, Maffei2015FastMC}, we implement this intuition by defining a prior spatial probability distribution for the user’s location, which in our case is derived from a measure of how well the observed walls match with local characteristics of the floor plan. Then, the customary particle evolution process is started, driven by the sensor data recorded as the user is walking. Particles that hit a wall are resampled from the remaining particles, which typically leads to a rapid coalescing of the particle into one cluster.  The distribution of particles is tracked until convergence is declared, at which time the user’s location is assumed to be reliably determined.

Here are the main contributions of this work:
\begin{itemize}
    \item We present a novel method (PALMS) to assess the probability that a 3-D scan, captured at a location using an iPhone equipped with LiDAR and processed to extract planar walls, was taken at a certain location in a floor plan for one of N possible orientations of the LiDAR's reference frame (where in our case, $N=4$). This probability is computed on the basis of the wall patches extracted from the LiDAR scan, which are then matched against the walls marked in the floor plan. A probability distribution of locations across the floor plan ({\em heatmap}) is quickly computed through by means of a linear convolution   for each possible orientation of the LiDAR's reference frame.
    
    \item We propose a method for dynamic localization that uses a particle filter initialized from the heatmap computed at each of the $N$ considered orientations

    \item We introduce a 2-step criterion  that can be used to declare ``convergence", the point in time after which the current user's location can be assumed to be correctly identified. Our experimental data shows that the proposed PALMS-based initialization leads to reduced convergence time and increased localization accuracy after convergence compared to a uniform initial probability distribution. In particular, the average RMSE for PALMS was found to be 6.7 times less than for \textit{Uniform}, while the proportions of locations with error less than 1 meter was found to be 4.9 times higher for PALMS than for \textit{Uniform}.
    
\end{itemize}

\begin{figure*}[t]
    \centering
    \includegraphics[width=\textwidth]{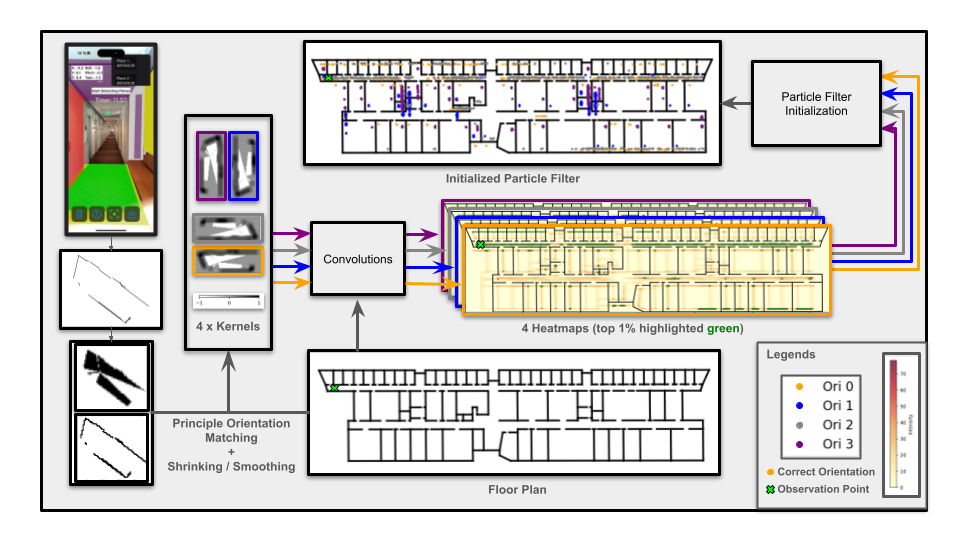}
    \caption{Architecture of PALMS' particle filter initialization pipeline. The ARKit-based application (top left) detects vertical planar patches (walls) in a $360^{\circ}$ LiDAR scan. We used their 3-D poses to create 4 convolution kernels through principal orientation matching, shrinking, or smoothing. Each kernel is rotated to match a possible orientation. We then convolve the rasterized floor plan with each of the kernels to create 4 heatmaps $H_{0\sim3}$. We extract the top 1\% locations from each heatmap and randomly sample 4 groups of particles, each with $p$ particles sharing the same initial drift from the corresponding orientation. These particles will then be placed in one particle filter. Colored edges and dots indicate different orientation groups. Orange represents the correct orientation, and the green cross indicates the ground truth observation point.}
    \label{fig:PALMS}
\end{figure*}

\section{Related Work}
Smartphone-based pedestrian tracking and localization systems in indoor, GPS-denied environments have received considerable attention recently. 
Many researchers have proposed schemes to estimate the user's location based on the Received Signal Strength Indicator (RSSI) values of wireless technologies, including WiFi \cite{Ito2014WRGBDFI}, Bluetooth Low Energy (BLE) beacons, Radio Frequency Identification (RFID), Ultra-Wideband (UWB), and magnetic field fingerprinting \cite{Tsai2024AllTW}. W-RGB-D \cite{Ito2014WRGBDFI} uses a heatmap of WiFi signal strength to initialize a particle filter for quick convergence in ambiguous environments.
We follow a similar particle filter initialization approach as described in section \ref{subsection:heatmap}. However, RSSI methods' requirement for prior calibration or "fingerprinting" before use restricts the scalability of these methods. Similarly, image-retrieval methods \cite{Dai2023EAAINetAE} require a pre-built image database, while 3-D structure-based methods \cite{Zhou2018ImageBasedLA} require pre-built SfM models. 

In contrast to fingerprinting-based approaches, some prior works perform indoor localization without prior visitation, leveraging only a known floor plan. 
Among these works, LiDAR-based methods are particularly notable.
Such methods utilize 3-D-to-2-D point cloud matching \cite{Watanabe2020RobustLW}, geometrical feature matching \cite{Wang2019GLFPGL}, usually coupled with particle filters \cite{Li2022LocalizationOI} or graph optimizations \cite{Gao2022FPLocLA}.
On the other hand, image-based methods \cite{Crabb2023ALA, Zhou2018ImageBasedLA, Goswami2023EfficientRL, Chen2019MultilevelSM, Yoo2022AVI, MendezMaldonado2017SeDARS, Zimmerman2022RobustOL, Chu2015YouAH, Hile2008PositioningAO, HowardJenkins2021LaLaLocLL, HowardJenkins2022LaLaLocGF} aim to perform localization from single or multiple images by utilizing image features \cite{Chen2019MultilevelSM}, semantic information (doors \cite{Goswami2023EfficientRL, Yoo2022AVI, MendezMaldonado2017SeDARS}, windows \cite{MendezMaldonado2017SeDARS}, texts \cite{Zimmerman2022RobustOL}, and exit signs \cite{Crabb2023ALA}), or scene geometries (vertical lines and vertical planes \cite{Chen2019MultilevelSM}). \cite{Chu2015YouAH} samples 3-D points from a raised 2-D floor plan and performs point cloud matching with the RGB-reconstructed 3-D scene.

While these methods generally rely on a flow of information over multiple sequential sensor measurements, some approaches try to tackle indoor localization by considering only an instantaneous observation \cite{Yoo2022AVI, Hile2008PositioningAO}. More recently, LaLaLoc\cite{HowardJenkins2021LaLaLocLL} and LaLaLoc++\cite{HowardJenkins2022LaLaLocGF} used deep neural networks to find “embeddings” for panorama images to match with the “embeddings” of 2-D locations on the floor plans. As pointed out by most of these papers, the ambiguity introduced by layout-similarity cannot be completely removed when localizing with respect to layout alone, making it hard to localize using a single observation.

In this work, we consider a dynamic form of finger-printing-free localization that utilizes a publicly available floor plan, an instantaneous observation, and a particle filter updated by odometry data (see section \ref{sec:method}).

\section{Method}
\label{sec:method}

\subsection{Plane Detection and Principal Orientation Matching}\label{sec:PDPOM}

We detect vertical planar patches (walls) in a $360^{\circ}$ LiDAR scan and obtain their 3-D poses using a custom ARKit-based iOS app. 
These patches are then projected onto the ground plane, forming a set of 2-D line segments (in metric units) $\mathcal{L}_{\text{obs}} = \{l_{obs, 1}, l_{obs, 2}, ..., l_{obs, n} \}$. We also represent the walls in the given floor plan as a set of 2-D line segments $\mathcal{L}_{\text{fp}} = \{l_{fp, 1}, l_{fp, 2}, ..., l_{fp, m} \}$, also expressed in metric units.

Intuitively, to localize the {\em observation point} (the center of projection of the LiDAR scan), we could try place \( \mathcal{L}_{\text{obs}} \) over \( \mathcal{L}_{\text{fp}} \) while adjusting \( \mathcal{L}_{\text{obs}} \)’s orientation and location, as if placing down a piece of a jigsaw puzzle looking for a match.
We limit the set of orientations by observing that, in most buildings, only a discrete set of wall orientations is observed. In our experiments, we considered buildings with walls intersecting at $90^{\circ}$, resulting in 2 possible orientations for any wall. Note that our algorithm can be easily extended to more complex buildings (e.g., walls intersecting at a multiple of $45^{\circ}$).

Our first task is to determine an angle $\theta$ such that, by rotating all observed line segments $\mathcal{L}_{\text{obs}}$ by $\theta$, any one such segment is aligned with one or more line segments in the floor plan. Note that if $\theta$ satisfies this property, all angles $\theta + k \cdot 90^{\circ}$ for $0\leq k<3$ will also satisfy it, and thus need to be considered. We determine $\theta$ by first finding the two orthogonal principal orientations of the observed line segments as well as of the floor plan segments. $\theta$ is taken to be the smallest angle that pairwise aligns the principal orientations found in the two domains.

\subsection{Heatmaps Creation}
\label{subsection:heatmap}

Consider the observed line segments $\mathcal{L}_{\text{obs}}$, rotated by $\theta + k \cdot 90^{\circ}$. Our goal is to define, for any point in the floor plan, the likelihood that this point represents the observation point. Ideally, at the correct location and orientation, the observed and floor plan line segments should overlap. By rasterizing these segments into bitmaps and multiplying them pixel by pixel, we obtain a score that, after normalization, defines a probability distribution for the location. Testing multiple locations is akin to computing a convolution on the rasterized floor plan, with the convolution kernel (termed $RW$ for recorded walls) representing the walls observed in the LiDAR scan (see Fig.~\ref{fig:PALMS})

\begin{figure}[t]
    \centering
    \includegraphics[width=\columnwidth]{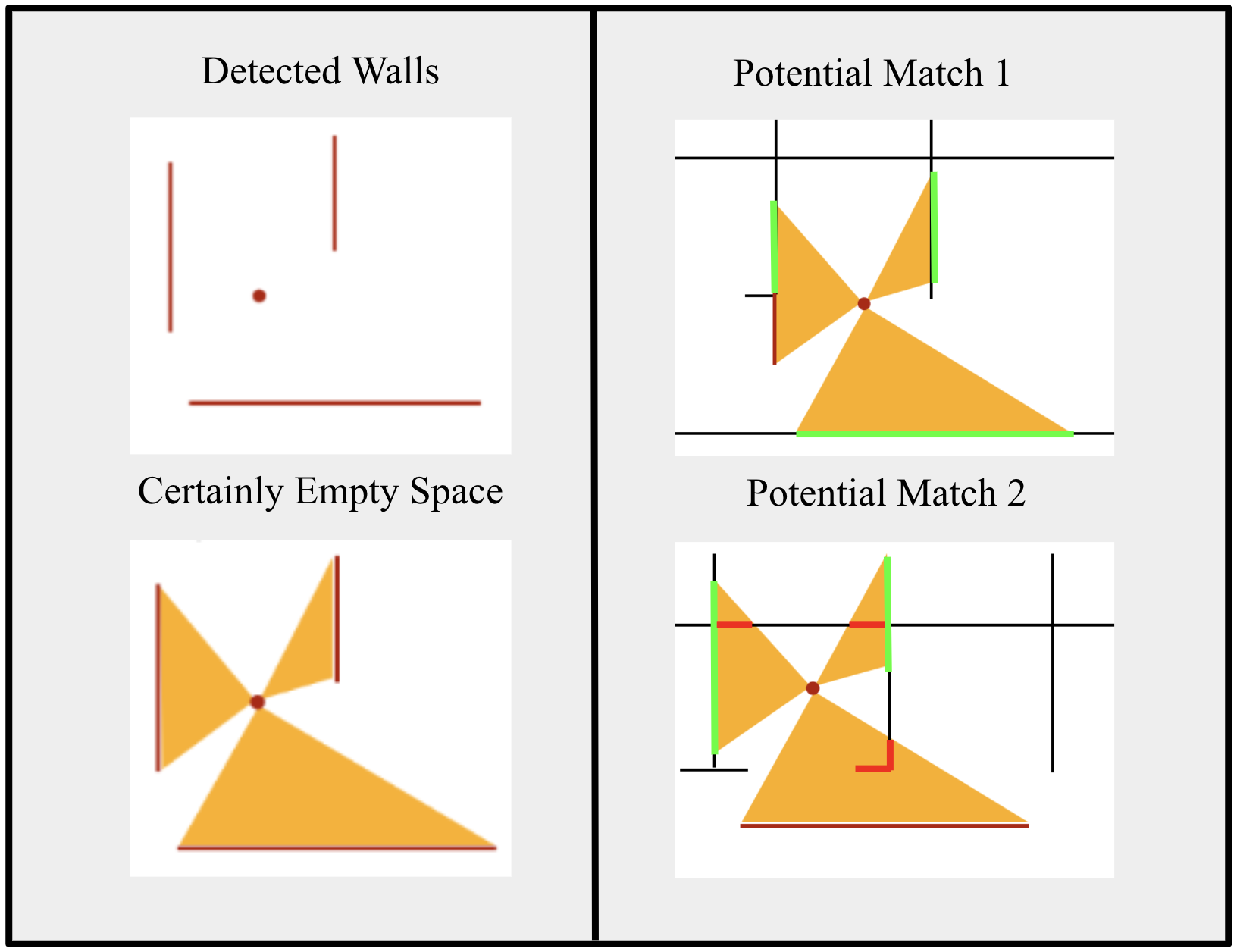}
    \caption{Certainly Empty Space (CES). CES enforces visibility constraints to eliminate physically impossible matches. The figure shows two scenarios: in match 1 (top right), the constraint is met as no floor plan segments fall within the CES (orange). In match 2 (bottom right), several segments intersect with the CES (bright red), making it a less viable match.}
    \label{fig:CES}
\end{figure}

\begin{figure*}[t]
    \centering
    \includegraphics[width=\textwidth]{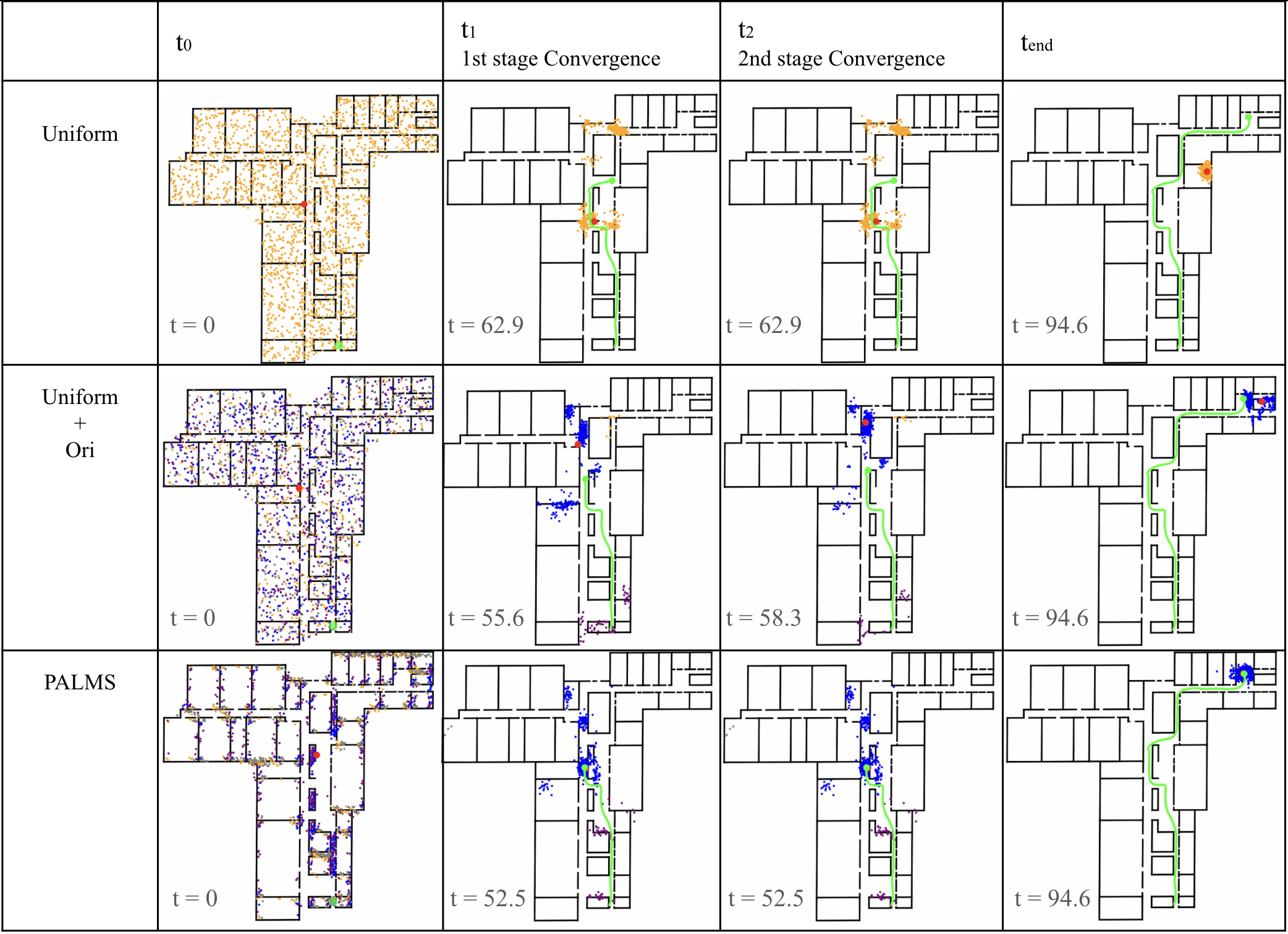}
    \caption{Visualization of the particle filters under different experiment settings at the time of initialization, at different stages of convergence, and when the path ends. Red dot: mean location of all particles (before $t_1$), mean position of the dominant group (after $t_1$), mean position of the dominant cluster (after $t_2$). Green: ground truth path and location. All three settings share the same starting point, tracking data, ground truth path, and particle count. Particles in different orientation groups are represented by different colors. 
    $t = x$ shows the time in seconds. We can see that PALMS successfully localized the user at $t_2$ and tracks the user until $t_{end}$ while the other methods failed to do so. Note that for some cases the 1st stage and 2nd stage convergence happen simultaneously.}
    \label{fig:table}
\end{figure*}

While intuitively appealing, this simple approach suffers from two main drawbacks. First, both the floor plan and the pose of the walls extracted from the LiDAR scan may be affected by errors. As a result, the observed line segments may not perfectly overlap with the floor plan segments even when $\mathcal{L}_{\text{obs}}$ is centered at the correct observation point. To account for this, we smooth the observed line segments bitmap with a Gaussian kernel, which basically ``widens" the line segments to account for possible localization errors. Second,  this criterion does not leverage visibility constraints.  Suppose that an observed line segment is matched to a floor plan segment and that another floor plan segment is located between the location being inspected and this matched segment (see Fig.~\ref{fig:CES}). This unmatched floor plan segment would not be penalized by our simple ``goodness of fit" measure, yet its presence is physically impossible, as it would occlude view of the matched wall segment. 

In order to enforce visibility constraints, we introduce the notion of {\em certainly empty space (CES)} (see Fig.~\ref{fig:CES}), which represents the areas where, based on the observation from a certain location, no wall in the floor plan should be found. Specifically,  if a wall (line segment) was observed from a certain location, no other floor plan line segment should be found in the triangle subtended by the observed line segment at that location. The union of such triangles for all visible line segment forms the CES area. By centering the CES space on different locations in the floor plan, we can find which locations break the visibility constraint (because they contain walls within the CES area). This is implemented by defining a second   kernel ($CES$), with values of 1 at the CES area and 0 otherwise, and convolving this with the rasterized floor plan. To account for possible wall localization errors, we first slightly ``shrink" (rescale) the $CES$ kernel, then combine the two kernels into one, equal to $RW - \alpha \cdot CES$, where we set $\alpha=1$ in our experiments. 

In practice, we first rotate the observed line segments $\mathcal{L}_{\text{obs}}$ by angle $\theta$ (Sec.~\ref{sec:PDPOM}), and compute the kernel $RW - \alpha \cdot CES$. Convolution of the rasterized floor plan with this kernel forms a {\em heatmap} $H_0$. We then rotate the kernel by multiples of $90^{\circ}$ to compute additional heatmaps $H_1,H_2,H_3$. The heatmap $H_k$ represents the probability distribution of the observation point across the floor plan, given that the reference frame for the LiDAR scan was at an angle of $\theta + k\cdot 90^{\circ}$ with the floor plan's reference frame. 

\subsection{Particle Filter Initialization from Heatmaps}
In our experiments (Sec.~\ref{sec: exp}), the ``ground truth" observation point \( P_{gt} \) consistently appeared among the locations with top \(1\%\) of all heatmap values. We thus simplify all heatmaps by binarizing them, using a threshold equal to the 99 percentile of all values in all maps (see Fig.~\ref{fig:PALMS}). Each location marked as 1 in a binarized heatmap $H_k$ represents a candidate observation point for the $k$-th orientation. Note that the proportion of 1's may vary across heatmaps for different orientations. 

We use these binary heatmaps to initialize a  particle filter tracking the motion of the user as they walk away from the observation point as driven by sensor data. Following~\cite{Ren2021SmartphoneBasedIO},  each particle tracks a state formed by location and angular drift. The filter is initialized as follows.  For each binary heatmap, we generate a number of particles sampled at uniform spatial density within the areas where the heatmap is equal to 1.  We then place all generated particles on the floor plan, where particles generated from the $H_k$ heatmap are assigned a label ($k$) and are given an initial orientation drift of  $\theta + k \cdot 90^{\circ}$. During  particle filter updating, particles that hit a wall are resampled from the remaining particles (any group), which typically leads to a rapid coalescing of the particles into one cluster. Note that when a particle is resampled from a particle with label $k$, the resampled particle maintains the same label, location (with noise added), and angular drift (with noise added).

\subsection{Convergence Criterion}
  It can thus be expected that the particle filter will converge to the correct user's location after a while (global localization problem~\cite{Dellaert1999MonteCL}). For example,~\cite{Crabb2023ALA} shows an empirical characterization of the time it takes for the location error to be less than 1 meter. However, in practical scenarios, the system does not have access to this error metric. Therefore, a measurable criterion is necessary to reliably declare "convergence", allowing the user to assume that their location is being tracked accurately from that point onward. We propose a 2-step convergence criterion based on the particles' distribution.

In \textbf{Step 1}, we check if one label dominates the current particle set (at least 80\%), since particles with the correct initial orientation are less likely to hit a wall and more likely to survive. We mark this point as $t_1$. 

In \textbf{Step 2}, we cluster particles with the dominant orientation using the mean shift algorithm and check if the dominant cluster contains at least 50\% of these particles. At this point $t_2$, we consider the particle filter fully converged, and the algorithm outputs the mean position of the dominant cluster as the predicted user location $P_{pred}$ (See Fig. \ref{fig:table}). Like \cite{Ren2021SmartphoneBasedIO}, we run the mean-shift algorithm to update the dominant cluster every 20 time steps after $t_2$ to keep track of the user's location.

\section{Experiments}
\label{sec: exp}

\subsection{Data Collection}
We collected data from 14 paths (2 per 7 starting points) in 3 buildings, averaging 147m in length. Each dataset includes LiDAR-detected vertical planes (using an iPhone 14 Pro) at the starting location and sensor data recorded along the paths. We scanned the surroundings by rotating the phone $360^\circ$ and walked away from the observation point, holding the phone with the camera facing forward to collect odometry data using iOS' ARKit. We used this odometry data and a particle filter initialized at the correct location and orientation to generate "ground truth" location data. The dataset can be accessed via the following link: \href{https://github.com/Head-inthe-Cloud/PALMS-Indoor-Localization/tree/main}{https://github.com/Head-inthe-Cloud/PALMS-Indoor-Localization/tree/main}

\subsection{Global Localization with PALMS}
In this experiment, we evaluate the advantages of PALMS initialization using 3-D scan data. We compare it against two methods without this data. In the first method ({\em Uniform}), particles are uniformly sampled across the floor plan with random initial angular drift between $0^{\circ}$ and $360^{\circ}$. The second method ({\em Uniform + Ori}) uniformly samples particle locations and assigns initial angular drifts $\theta + k \cdot 90^{\circ}$ for each of the $k$ groups ($0\leq k<3$). $\theta$ is calculated using the principal orientations of $\mathcal{L}_{\text{fp}}$ and the first few tracking points. For {\em Uniform}, convergence Step 1 is determined by setting a threshold on particle dispersion, while Step 2 monitors for dominant clusters like PALMS. All three methods use 2000 particles and run 100 trials per path, with random re-initialization for each trial. 

Measurements included: time (in seconds) and distance (in meters) until convergence is declared; RMSE after convergence (in meters); and proportion of predicted locations (after convergence) at less than 1 meter from the ground truth ({\em \%Error\textless1m}). 
Results in Tab.~\ref{tab1} (averaged over all trials) show that PALMS significantly outperforms the other methods considered. In particular, the average RMSE for PALMS was found to be 6.7 times less than for {\em Uniform} (5.6 times less than for {\em Uniform + Ori}), while  the proportions of locations with error less than 1 meter post-convergence was found to be 4.9 times higher for PALMS than for  {\em Uniform} (2.8 times higher than for {\em Uniform + Ori}). Examples of particles distributions for the three cases at different stages of convergence are shown in Fig.~\ref{fig:table}.

\begin{table}[h]
\caption{Post Convergence Metrics for Global Localization}
\begin{center}
\begin{tabular}{|c|c|c|c|c|}
\hline
\textbf{Init.}&\multicolumn{4}{|c|}{\textbf{Metrics}} \\
\cline{2-5} 
\textbf{Method} & \textbf{\textit{Time (s) $\downarrow$}} & \textbf{\textit{Dist. (m) $\downarrow$}} & \textbf{\textit{RMSE $\downarrow$}} & \textbf{\textit{\%Error\textless1m $\uparrow$}} \\
\hline
Uniform & 75.94 & 129.87 & 31.52 & 16\% \\
\hline
Uniform + Ori & 69.18 & 121.86 & 26.03 & 28\% \\
\hline
PALMS & \textbf{59.14} & \textbf{110.28} & \textbf{4.69} & \textbf{78\%} \\
\hline
\end{tabular}
\label{tab1}
\end{center}
\end{table}

\section{Discussion and Conclusions}
We presented an indoor localization algorithm using an iPhone LiDAR scan to acquire nearby wall geometry, then finding locations in a floor plan consistent with the observed wall geometry for possible camera orientations. This information filters out unlikely locations but cannot precisely pinpoint the user due to geometric ambiguities in self-similar layouts. To address this, we generate a spatial probability distribution for each orientation to initialize a particle filter, which is expected to converge to the correct location.

We showed that this strategy reduces time to convergence and substantially increases accuracy after convergence compared to uniform particle initialization (i.e., without initial orientation and the initial information on the user location provided by the LiDAR scan).
A main advantage of our approach is that it does not require prior “fingerprinting” of the environment, as it only uses a floor plan for prior information. It provides a solution to a critical bottleneck of dead-reckoning tracking algorithms (based on visual or inertial data), that is the determination of the initial user location and orientation. 

At the same time, there are a number of limitations that need to be considered. Our system uses an iPhone LiDAR, which is only available on “Pro” models at the time this paper is written. In addition, the maximum range of this sensor is 5 meters, which reduces its utility in wide open spaces. In future work, we will consider extracting planar wall models directly from images to enable detection of walls beyond the LiDAR’s range, although it could be less accurate. While heatmap-based particle initialization has been shown to reduce convergence time, this system still requires the user to walk for a while (up to a minute) before convergence. This could prove impractical, and we will consider strategies like combining other modalities to reduce convergence time in future work. Finally, the dead-reckoning tracker we considered in this work was based on visual-inertial odometry from data recorded by an iPhone held by the user while walking. We will consider applying the same strategy to inertial-only tracking with particle filtering. In this configuration, users can use the phone to take a LiDAR scan, then conveniently put the phone back in their pocket, and be tracked by the phone’s inertial sensors.

\section{Acknowledgements}
Research reported in this publication was supported by the National Eye Institute of the National Institutes of Health under award number R01EY029260-01. The content is solely the responsibility of the authors and does not necessarily represent the official views of the National Institutes of Health. 

\bibliographystyle{IEEEtran}

\bibliography{references} 
\end{document}